\documentclass{article} % For LaTeX2e
\usepackage{iclr2025_conference,times}

\usepackage{amsmath,amsfonts,bm}

\def\eqref#1{equation~\ref{#1}}
\def\1{\bm{1}}

\DeclareMathAlphabet{\mathsfit}{\encodingdefault}{\sfdefault}{m}{sl}
\SetMathAlphabet{\mathsfit}{bold}{\encodingdefault}{\sfdefault}{bx}{n}

\usepackage{hyperref}
\usepackage{url}

\title{\model~: Text and Frequency Guidance for Subject Driven Action Personalization using Diffusion Models}

\author{Divya Kothandaraman*,\And Kuldeep Kulkarni,\And Sumit Shekhar,\And Balaji Vasan Srinivasan, \And Dinesh Manocha* \\ \AND University of Maryland College Park*, \And Adobe Research}

\usepackage{amsfonts}       % blackboard math symbols
\usepackage{wrapfig,lipsum,booktabs}
\usepackage{algorithm,algpseudocode}
\usepackage{amssymb}
\usepackage{graphicx}
\newcommand{\model}{ImPoster}

\iclrfinalcopy % Uncomment for camera-ready version, but NOT for submission.
\begin{document}

\maketitle

\begin{abstract}
    We present ImPoster, a novel algorithm for generating a target image of a `source' subject performing a `driving' action. The inputs to our algorithm are a single pair of a source image with the subject that we wish to edit and a driving image with a subject of an arbitrary class performing the driving action, along with the text descriptions of the two images. Our approach is completely unsupervised and does not require any access to additional annotations like keypoints or pose. Our approach builds on a pretrained text-to-image latent diffusion model and learns the characteristics of the source and the driving image by finetuning the diffusion model for a small number of iterations. At inference time, ImPoster performs \emph{step-wise text prompting} i.e. it denoises by first moving in the direction of the image manifold corresponding to the driving image followed by the direction of the image manifold corresponding to the text description of the desired target image. We propose a novel diffusion guidance formulation, \emph{image frequency guidance}, to steer the generation towards the manifold of the source subject and the driving action at every step of the inference denoising. Our frequency guidance formulations are derived from the frequency domain properties of images. We extensively evaluate ImPoster on a diverse set of source-driving image pairs to demonstrate improvements over baselines. To the best of our knowledge, ImPoster is the first approach towards achieving both subject-driven as well as action-driven image personalization. Code and data is available at https://github.com/divyakraman/ImPosterDiffusion2024.
\end{abstract}

\begin{figure*}   
    \centering
    \includegraphics[scale=0.27]{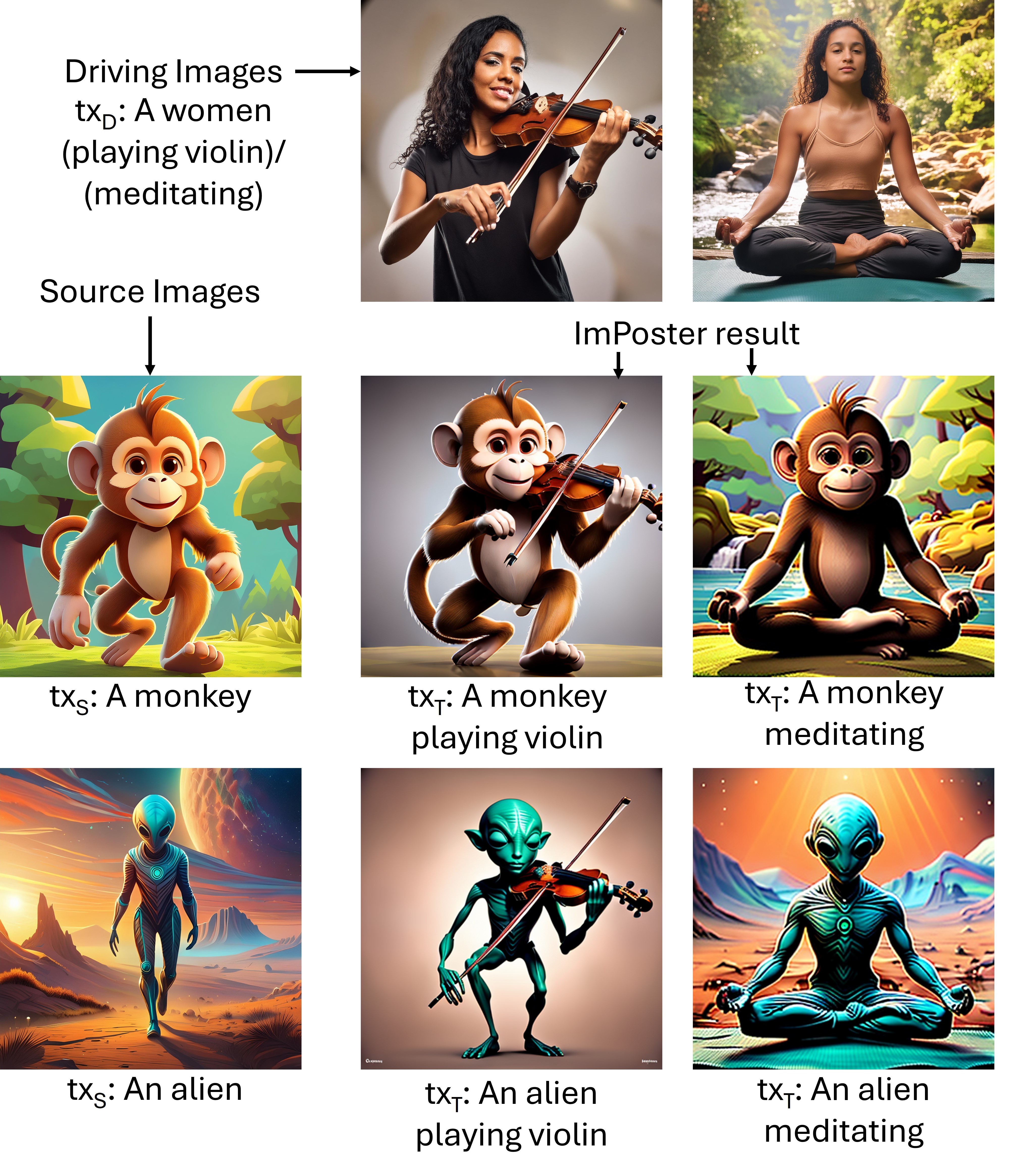}
    \caption{\textit{Given a single ``source'' image and a ``single'' driving image and the corresponding text descriptions, ImPoster generates an image of the source subject performing driving action.} We show how ImPoster is able to make a monkey and an alien meditate and play violin.}
    \label{fig:teaser}
    %\vspace{-10pt}
\end{figure*}

% Remove: SOTA table; Modify: User study    
\section{Introduction}

Monkeys seldom meditate! Neither do they play the violin! Generative AI has enabled the creation of images that cannot be easily enacted or photographed in real life. More often than not, applications such as animation, movie creation, etc, require a particular subject performing a specific action. For example, given a `source image' of a dog and a `driving image' of a cat drinking water from a mug, we might want to generate an image of the dog drinking water from a mug in exactly the same manner as the cat. Thus, the generated dog needs to have the same identity as the `source subject'. Its pose or the style with which it drinks water from the mug needs to resemble the `driving pose', i.e. the pose of the cat in the driving image. Consequently, the specific action depicted in the generated image and the identity of the subject performing the action are not arbitrary - rather, they are dictated by the source and the driving images. 

Prior work on the closely related problem of motion transfer~\cite{zhao2022thin,tao2022structure,siarohin2021motion,siarohin2019first,shalev2022image} condition on SMPL pose information~\cite{yoon2021pose}, keypoints, etc in a supervised/ unsupervised manner and train on large datasets to transfer the action depicted by a subject in a driving image or video to the subject in the source image. Such methods however do not generalize well to arbitrary pairs of source and driving images and often require a lot of data for training. Thus, it is useful to develop an approach that can directly work on a single pair of source-driving images in a completely unsupervised manner. 

Text is an excellent auxiliary modality to guide the generation process, and the recent progress in text-to-image diffusion models~\cite{rombach2022high,saharia2022palette,ho2022imagen} motivates their usage for conreolled image generation. Prior work on personalized text-guided image editing such as DreamBooth~\cite{ruiz2022dreambooth}, IMAGIC~\cite{kawar2022imagic}, custom diffusion~\cite{kumari2022multi}, InstantBooth~\cite{shi2023instantbooth}, \cite{gal2023designing}, ELITE~\cite{wei2023elite} and SUTI~\cite{chen2023subject} are able to add a specific subject to the model while finetuning the diffusion model to generate various actions dictated by the text input. However, the user does not have any control over the exact manner in which they might like the subject to perform the action in the generated image. For instance, a dog can drink water from a mug in various ways - we want the model to be able to generate a specific pose corresponding to the action, as defined by the driving image. On the other hand, prior work such as blended diffusion~\cite{avrahami2022blended}, DiffusionCLIP~\cite{kim2022diffusionclip} and ADI~\cite{huang2024learning} are able to generate a specific action and change the style of the image. However, they are unable to make a source subject perform a specific driving action. 

\paragraph{Main contributions.}
We propose an algorithm, \model, for generating an image of a specific subject from a source image performing a specific action described in the driving image. Given a single source-driving image pair and the corresponding text descriptions, our method can perform `body' transformations to the source subject, as dictated by the action depicted in the driving image. Our method builds on a pre-trained text-to-image diffusion model to perform test-time optimization/ inference and does not require any additional information such as pose or keypoints. Our method, \model~, first learns the characteristics of the source and the driving image by finetuning the pretrained diffusion model on the source and driving corresponding text-image pair. Our inference guidance methods, which form the novel contributions of this paper, enable the generation of the desired target image:
\begin{enumerate}
    \item \textbf{Stepwise text prompting.} \model~generates the desired target image of the source subject performing the specific driving action using a \emph{stepwise text prompting} strategy. Since geometric or structural information is much harder to generate, \model~lays the foundation for generating the driving action by first moving towards the direction of the manifold of the driving image. This step is followed by switching directions and denoising by moving in the direction of the text description corresponding to the desired target image. 
    \item \textbf{Image frequency guidance.} While the stepwise text prompting method provides a strong prior for the driving action, it is insufficient to generate an accurate image of the source subject. To alleviate this issue, we delve into the frequency domain representation of an image, a powerful space for understanding how the human brain interprets natural images~\cite{oppenheim1981importance,xu2021fourier,yang2020fda,yang2020phase}. Motivated by guidance techniques~\cite{ho2022classifier,bansal2023universal} in diffusion models, we harness the amplitude map of the Fourier transform of image features and present frequency amplitude guidance to preserve the characteristics of the source subject. In order to prevent the loss of the specific driving action (or structure) while extracting source subject characteristics, frequency amplitude guidance is complemented by frequency phase guidance. This is inspired by the fact that the phase of the Fourier transform of an image is representative of its geometry and structure. Frequency amplitude guidance and frequency phase guidance, termed as \emph{frequency guidance}, guide the diffusion denoising towards generating the source subject performing the specific driving action. 
\end{enumerate}

We apply \model~on a wide variety of source-driving image pairs on a curated dataset with $120$ source-driving image pairs. \model~ can make an elephant read a book (described by a human reading a book), a monkey meditate and perform push-ups (described by a human performing meditating and doing pushups), a teddy bear play guitar (described by a human playing guitar), etc. We also show the effectiveness of stepwise inference and our frequency guidance method, along with qualitative and quantitative comparisons against prior work metris such as CLIP Score, SSCD, DINO and a new metric to quantify the alignment with driving action, phase score, defined using the phase of the Fourier transform. 

In summary, the contributions of this paper are as follows: (i) We formalize a novel task of generating images consistent with a source subject and driving action; (ii) We propose ImPoster, a novel diffusion models based method that can address this pragmatic task; (iii) We curate a dataset of $120$ source-driving image pairs; (iv) In order to quantitatively establish the correspondence between the generated image and driving action, we propose a new metric `Phase score' based on the phase of the Fourier transform; (v) Finally, our exhaustive experimental results reveal large gains over baselines; thereby setting a new benchmark for the task.
\begin{figure*}
    \centering
    \includegraphics[scale=0.24]{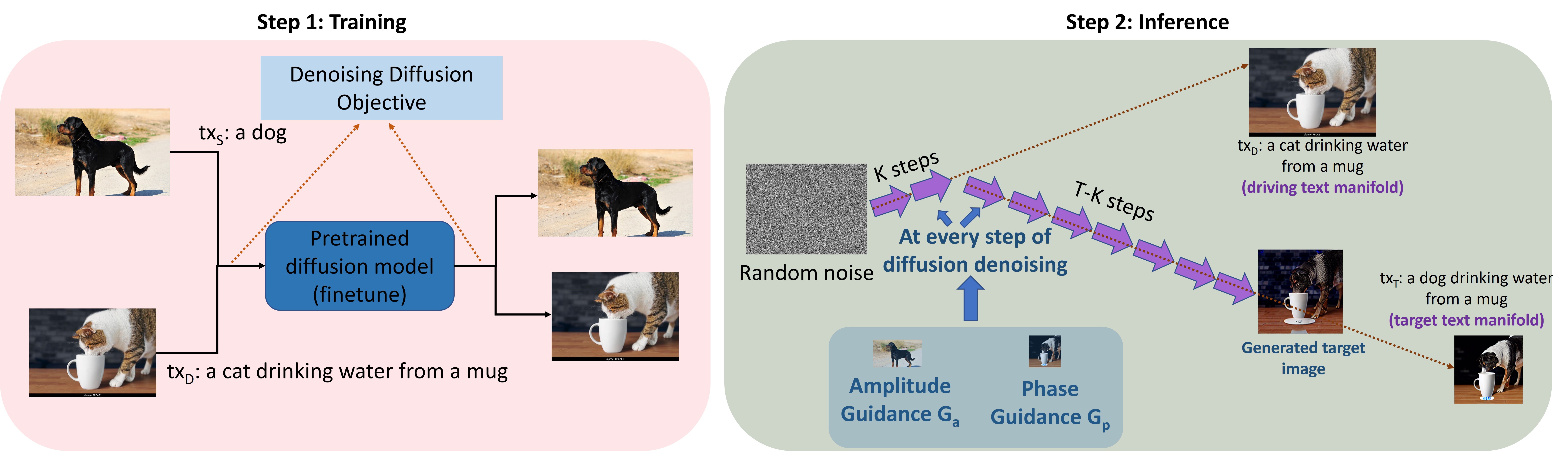}
    \caption{Given a single source-driving image pair, \model~generates an image of the source subject performing the action depicted in the driving image. \model~first finetunes the text-to-image diffusion model on the source-driving image pair. At inference, \model~begins by first denoising in the direction of the driving image manifold followed by moving towards the manifold corresponding to the desired target image. At every step of the inference, frequency guidance steers the generation of an image with source subject characteristics and driving action.}
    \label{fig:overview}
\end{figure*}

\section{Related work}

\paragraph{Exemplar image animation.} A great amount of literature has been dedicated to exemplar image animation~\cite{shalev2022image,siarohin2021motion,siarohin2019first,tao2022structure,zhao2022thin,siarohin2019animating,siarohin2018deformable,tao2022structure} transfer motion characteristics from a driving video to a source image. These methods at the core of them transfer motion at frame level from a driving frame onto the source image, similar to our method. However, these models are restricted in their domain to only human and almost always require additional annotations like keypoints, 2D poses, 3D poses or require computation of optical flow. %Our method is different in two ways, it can transfer imitable characteristics like pose from various driving subjects like dolphin, cat, human, rabbit, cat to other object categories like teddy bear, jerry mouse, dog, monkey, toy fireman. Secondly, our method does not require any additional annotations other than easily available or constructable text prompts. 

\paragraph{Diffusion models for Text-based image editing/ personalization.} Recent progress in generative AI has shown promising results using diffusion models~\cite{nichol2021glide,wang2022pretraining,su2022dual,sasaki2021unit,saharia2022palette,saharia2022image,yang2022paint,preechakul2022diffusion,zhang2023adding} for performing non-trivial operations, such as posture changes and multiple objects editing. Text-based image editing and personalization approaches~\cite{ruiz2022dreambooth,kawar2022imagic,kumari2022multi,balaji2022ediffi,gal2023designing,kothandaraman2023aerial,huang2023composer,brooks2022instructpix2pix,kothandaraman2024prompt,zheng2022bridging,shi2023instantbooth,zhang2022sine,ma2023unified,gal2022image, shi2023instantbooth, wei2023elite, chen2023subject, kumari2022multi,han2023svdiff,qiu2023controlling,ma2023subject,xiao2023fastcomposer,zhao2023dreamdistribution,mou2024t2i,huang2024learning} add a subject to the diffusion using a few images of the subject followed by using text to manipulate the image to obtain the desired output. The methods for personalized or subject-driven based image editing can be categorized into two broad categories. First category of works \cite{ruiz2022dreambooth,kawar2022imagic,wei2023elite,kumari2022multi,gal2023designing,yang2022paint} perform fine-tune a pre-trained model on a small number of images that involve the subject and then perform inference optimization with the text depicting the action as the input. While these methods are successful in generating images with the subject performing an action depicted by the text, the user does not have control over the specific pose or imitable characteristic. %We show through our comparison unlike the above methods later owing to the frequency based optimization in the latent space during inference time, o
%Our approach is able to not only retain the identity of the subject but also performs the action depicted by the text in a similar pose as the driving image. 
The second category of works \cite{shi2023instantbooth, chen2023subject} address the shortcomings of fine-tuning the model and propose methods that are purely based on inference-time optimization and hence much faster. However, even they suffer from the same drawbacks in not being able to give users the control of the action depicted. Multi-concept customization methods~\cite{kumari2022multi,ma2023subject,xiao2023fastcomposer,han2023svdiff} are not very effective either. For instance, one of the closest works to ours is Custom Diffusion \cite{kumari2022multi} that enable generating images by fusing multiple concepts from multiple images. However, this method is not well suited to generate images involving concepts of actions. Action customization methods~\cite{huang2024learning} emphasize more on action customization, and is not as effective when it comes to subject + action customization.

Recent methods on motion customization~\cite{wu2023tune,kothandaraman2024text,chen2023videodreamer} of diffusion models are able to transfer the motion from a video to subjects. However, these methods rely on temporal properties of videos to tranfer the motion. Our problem statement is tangential to video motion customization, wherein the goal is to customize the action from an image. 

\paragraph{Guidance methods for diffusion.} Guidance methods~\cite{dhariwal2021diffusion,ho2022classifier,kawar2022denoising,wang2022zero,chung2022diffusion,lugmayr2022repaint,chung2022improving,graikos2022diffusion,kothandaraman2023aerialbooth} have been used to control and guide diffusion denoising. One of the first guidance methods is classifier and classifier-free guidance~\cite{dhariwal2021diffusion,ho2022classifier} that reinforce the class of the object in the generated image. Bansal et. al.~\cite{bansal2023universal} proposed universal guidance to guide the generation using segmentation maps, sketches, etc. Kothandaraman et. al.~\cite{kothandaraman2023aerialbooth} proposed a mutual information based guidance method to generate high fidelity images from various viewpoint. %In this paper, inspired by classifier-free guidance, we propose a frequency guidance technique to impose the diffusion denoising to stay on the manifold of the source subject and the driving action. Our guidance formulation is simplistic and directly modifies the latent space using a suitable frequency formulation. 

\section{Method}

Given a source image $I_{S}$ and a driving image $I_{D}$, \model~generates an image of the source subject performing the driving action. We assume access to the text descriptions $tx_{S}$ and $tx_{D}$ for the source and driving images respectively. The text description, $tx_{T}$, corresponding to the desired target image $I_{T}$ is a modified combination of $tx_{S}$ and $tx_{D}$. For example, if $tx_{S}$ is ``A dog'' and $tx_{D}$ is ``A cat drinking water from a mug'', $tx_{T}$ would be ``A dog drinking water from a mug''. We assume no access to any training dataset, pose information (such as SMPL), keypoints, etc. Our method is completely unsupervised and works on a single source-driving image pair. 

An overview of \model~is as follows. To add knowledge pertaining to the source and the driving image, we finetune the diffusion model on $I_{S}$ and $I_{D}$ using $tx_{S}$ and $tx_{D}$, respectively. During inferencing, we begin by denoising by first moving in the direction of the manifold corresponding to $I_{D}$ followed by moving in the direction of the manifold corresponding to $tx_{T}$. Such a mechanism, which we term \textbf{stepwise text prompting} creates a prior for pose information followed by denoising to generate the desired subject performing the action. Further, to reinforce the generation of an image of the source subject performing the driving action, at every inference step, we apply a novel \textbf{image frequency guidance} strategy to explicitly steer the denoising towards the desired driving action and source subject characteristics. We now turn to describe our method in detail.

\subsection{Training}

A pretrained text-to-image diffusion model such as stable diffusion is trained on massive text-image data. However, it does not contain information specific to the source and driving image. In order to enable the diffusion model reconstruct $I_{S}$ and $I_{D}$ from $tx_{S}$ and $tx_{D}$, respectively, we finetune the diffusion model for $n_{tr}$ iterations on the source and driving text-image pairs using the denoising diffusion objective function~\cite{ho2020denoising}. Specifically, let $\theta$ denote the U-net parameters, $L$ denote the denoising diffusion objective, and $e_{S}$, $e{D}$ denote the text embeddings corresponding to the source and driving texts. At each iteration $i < n_{tr}$ of finetuning, we perform the following optimizations:
\begin{equation}
\begin{split}
        \min_{\theta} \sum_{t=T}^0 L(f(x_t, t, e_{S};\theta), I_{S}), \\ \min_{\theta} \sum_{t=T}^0 L(f(x_t, t, e_{D};\theta), I_{D}).
\end{split}
\end{equation}

\subsection{Inference}

\subsubsection{Stepwise text prompting}

A key question that we need to keep in mind while developing an effective inference strategy is: \emph{How should the noisy image be perturbed so that it better corresponds with the driving action and the source subject ?} Moving in the direction of the driving image manifold will ensure the reconstruction of the driving action accurately, however, biased towards the driving subject. Similarly, moving in the direction of the source image manifold will ensure the reconstruction of the source subject accurately, however, biased towards the source action. The image manifold corresponding to $tx_{T}$ generally does not contain the desired target image -- i.e. given $tx_{T}$, it is unable to generate the desired target image by naive diffusion denoising. On one hand, a pretrained diffusion model conditioned on $tx_{T}$ is inclined to generate a wide variety of images, most not fully conformant with the specific characteristics of the source subject as well as the specific pose delineated by the driving action. On the other hand, there could be a bias towards reconstructing the source or driving image as well. 

To solve the aforementioned issues, we propose a stepwise inference strategy. Let the number of inference steps be denoted by $T_{i}$. Starting from random noise, we denoise using $tx_{D}$ for the first $K$ iterations following by denoising using $tx_{T}$ for the next $T-K$ iterations. This implies that the generation process begins by first moving towards the driving image manifold followed by the manifold corresponding to the target text. In the first few iterations, the sampler denoises to reconstruct the driving image, creating a strong prior for the driving action. In the subsequent iterations, denosing to move towards the target text manifold ensures that the generated image contains the source subject, as dictated by the target text. 

On one hand, the target text points towards the driving action. On the other, denoising using $tx_{D}$ in the first few iterations creates a strong prior for the structure of the generated image. This ensures that the generated image has the driving action. Note that we denoise with $tx_{D}$ first followed by $tx_{T}$ and vice versa. Diffusion works by sequentially denoising from random noise and is a slow process i.e. the updates in each step are small. Therefore, it is easier for the model to first generate the pose and then denoise to obtain subject characteristics. It is much more difficult to modify the pose than it is to modify the characteristics or identity of the subject in the image.

Despite the stepwise inference strategy, the generated image might still contain certain characteristics of the driving subject or source action. To alleviate this issue, we propose a novel frequency guidance method, which reinforces the model to stick to the driving action and source subject at every step of the generation process. 

\subsubsection{Frequency guidance}

The frequency domain representation of an image provides rich information~\cite{oppenheim1981importance} about the image. The amplitude of the 2D spatial Fourier transform of an image is representative of the intensities of different frequencies in the image, and represents changes in the spatial domain. It contains information about the geometrical structure of features in the image. The phase of the 2D spatial Fourier transform of an image represents the location of these features which help the human eye understand the image; it contains information regarding the edges, contours, etc. It is possible to reconstruct the grayscale counterpart of an image with just its phase representation. The amplitude information, along with the phase, helps in reconstructing the characteristics (colors, attributes, texture, identity) of the scene entities. However, the amplitude alone cannot generate a meaningful image; phase information is very important.

In light of this, it can be inferred that the phase of the Fourier transform of the driving image is indicative of the driving action. Similarly, the amplitude of the Fourier transform of the source image is indicative of the knowledge required to reconstruct the source image. We use the Fourier domain representation to \textbf{guide} the generation process. Frequency domain properties of an image propagate to its feature space (computed by a neural network). Hence, we can apply frequency guidance on the latent feature representation space of stable diffusion~\cite{rombach2022high}. Classifier-free guidance~\cite{ho2022classifier} and universal guidance~\cite{bansal2023universal} mathematically derive guidance as,
\begin{equation}
    \tilde{\epsilon_{\theta}} (z_{t}, t) = \epsilon_{\theta} (z_{t}, t) + s(t) \times \nabla_{z_{t}}G.
\end{equation}
where $z_{t}$ is the denoised latents at timestep $t$, $\nabla_{z_{t}}$G is the gradient of the guidance function and $s(t)$ controls the strength of the guidance for each sampling step. An appropriately weighted version of this noise is subtracted from the latents computed in the previous timestep to obtained updated latents. Consequentially, rather than adjusting the noise predicted by the diffusion model using the gradients of the guidance function, we can directly modify the computed latents using a suitable guidance function using the gradient of the computed amplitude and phase functions w.r.t. the latents. At every step of the inference sampling process, we modify the computed latents $z_{t}$ as,
\begin{equation}
    \tilde{z_{t}} = z_{t} - s_{a} \times \nabla_{z_{t}}G_{a} - s_{p} * \times \nabla_{z_{t}}G_{p},
\end{equation}
where $s_{a}$ and $s_{p}$ are the scaling factors for the frequency amplitude guidance and frequency phase guidance functions, $G_{a}$ and $G_{p}$, respectively. Note that $z_{t}$ is computed using the noise predicted by the diffusion model at each timestep. $G_{a}$ is the $L2$ distance between the amplitude of the generated latents and the amplitude of the latents of the source image. $G_{p}$ is the $L2$ distance between the phase of the generated latents and the phase of the latents of the driving image.
%\begin{equation}
%    G_{a} = i\mathcal{F}(Mag(\mathcal{F}(f_{s}))), G_{p} = i\mathcal{F}(Ang(\mathcal{F}(z_{t}))),
%\end{equation}
%where $f_{s}$ is the latent space representation of the source image, $i\mathcal{F}$ and $\mathcal{F}$ denote the operations of inverse FFT and FFT respectively; Mag and Ang denote the magnitude and the angle of the Fourier transform. 
$G_{a}$ drives the generation towards the source subject at every step of the sampling. $G_{p}$ reinforces the driving pose at every step of the sampling and prevents any distortion in pose that may be caused by $G_{a}$. 

\begin{figure*}[h]
    \centering
    \includegraphics[scale=0.45]{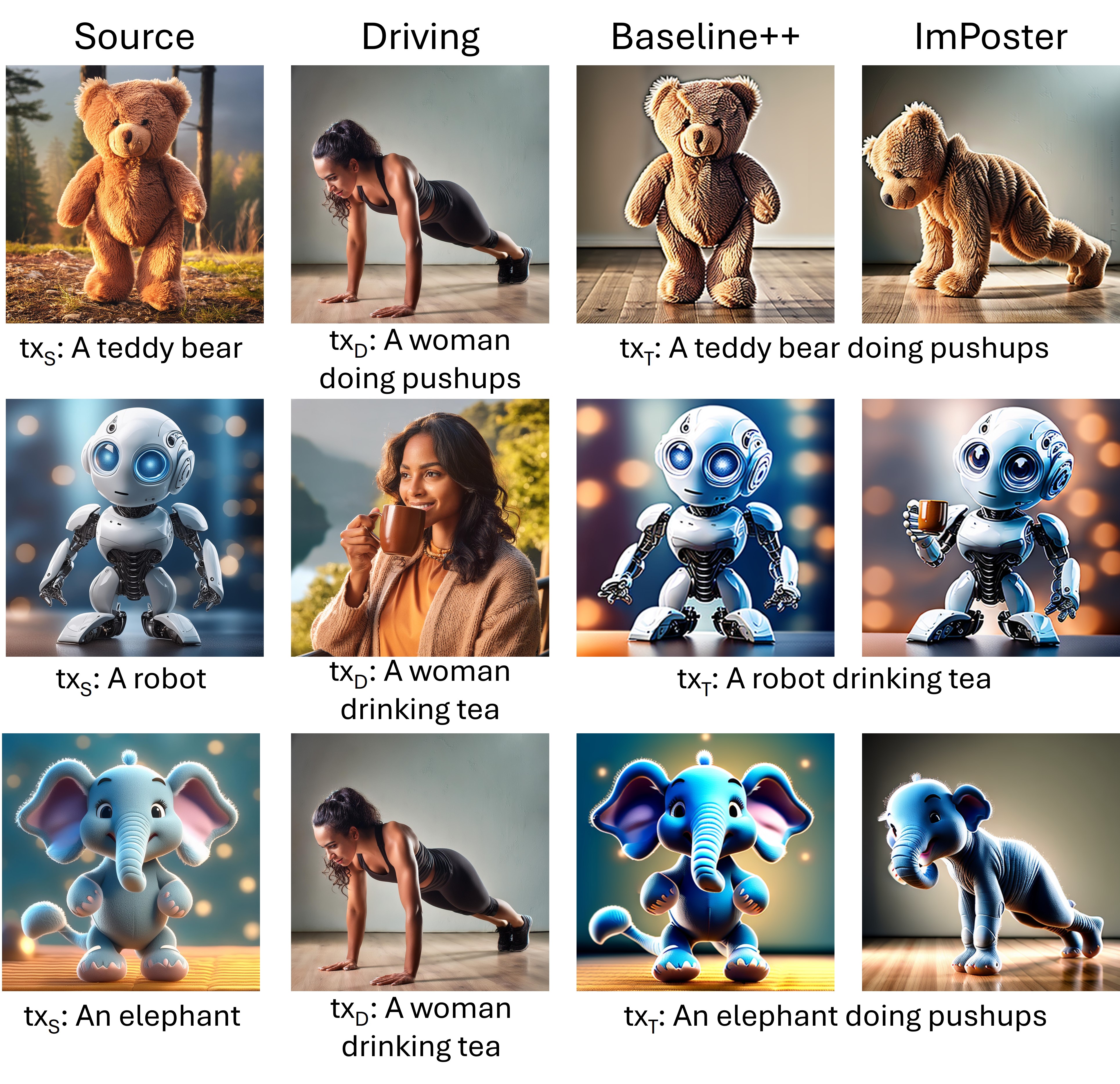}
    \caption{ImPoster is able to successfully transfer the driving action to a source subject, while maintaining its characteristics. In contrast, Baseline++ is unable to generate the driving action for the given source subject due to bias issues. Please see the appendix for more results generated using ImPoster and comparisons with Baseline++.}
    \label{fig:results1}
    %\vspace{-25pt}
\end{figure*}

\section{Experiments and Results}

\paragraph{Dataset and implementation details.} Since there is no dataset for this task, we collect a dataset with $15$ driving actions and $8$ source subject, resulting in a total of 120 source-driving image pairs. The source subjects and driving actions are generated using Adobe Firefly. This dataset curation is inspired by prior papers that defined new tasks in diffusion personalization such as DreamBooth~\cite{ruiz2022dreambooth}(30 subjects), Custom Diffusion~\cite{kumari2022multi}(5 concepts, 8 prompts for the multi-concept setting), and Concept Decomposition~\cite{vinker2023concept}(15 pairs for application 1, 13 concepts for application 2) and is also comparable in size to these prior datasets. Consistent with prior work, we generate results corresponding to $5$ different random seeds for each source-driving pair, resulting in a total of $600$ images being used for all quantitative analysis. 

We finetune the diffusion model for $n_{tr} = 500$ iterations. We set $s_{a}$ to $1e-6$, $s_{p}=1e-3$, $k=5$ and $T=50$. We use an image size of $512\times512$. Our model takes about $5 1/2$ minutes per image pair on one NVIDIA RTX A5000 GPU. We use the Stable Diffusion 2.1 model as the backbone. We add LoRA~\cite{hu2021lora} layers to the backbone model for the finetuning step, the rest of the model is frozen. 

\paragraph{State-of-the-art comparisons.} Closest related to our work are image editing or personalization approaches. The most recent and effective methods in this domain are DreamBooth~\cite{ruiz2022dreambooth}(CVPR 2023) and Custom Diffusion~\cite{kumari2022multi}(CVPR 2023), inspired by which, we compare with an enhanced baseline, Baseline++. Baseline++ takes in a single source image and driving image along with their corresponding text prompts and finetunes the diffusion model (similar to DreamBooth~\cite{ruiz2022dreambooth} + LoRA~\cite{hu2021lora}). Next, it uses a pragmatic combination of the text prompts corresponding to the source and the driving image ($tx_{T}$) to generate the target image (similar to Custom Diffusion~\cite{kumari2022multi}). 

We show comparisons with Baseline++ in Figure~\ref{fig:results1}. Baseline++, due to bias issues with respect to the pose in the source image, and inability to effectively generate the driving action, is unable to generate the source subject performing driving action. In contrast, ImPoster holistically distils the driving action and the characteristics of the source subject through its stepwise text prompting and image frequency guidance strategies to achieve a good bias-variance trade-off and generate the source subject performing driving action. 
%In the first example, the leg of the jerry mouse is not lifted up in the result corresponding to Baseline++. ImPoster fixes the pose of the jerry mouse. In the second result of Baseline++, the identity of the dog is lost and the size of the dog is slightly different, but is better retained by ImPoster. In the third example, both pose and attribute are lost in Baseline++'s result - the wrong hand is lifted up, one hand is missing, the color of the clothes is slightly white and the princess is marginally turned to the front rather than the side. All of these issues are resolved by ImPoster. In the fourth case, the specific pose corresponding to the action is completely lost. ImPoster is able to generating the specific action, while preserving the identity of the source subject to a reasonable extent. 

%\input{Figures/table2}

\begin{figure*}[h]
    \centering
    \includegraphics[scale=0.45]{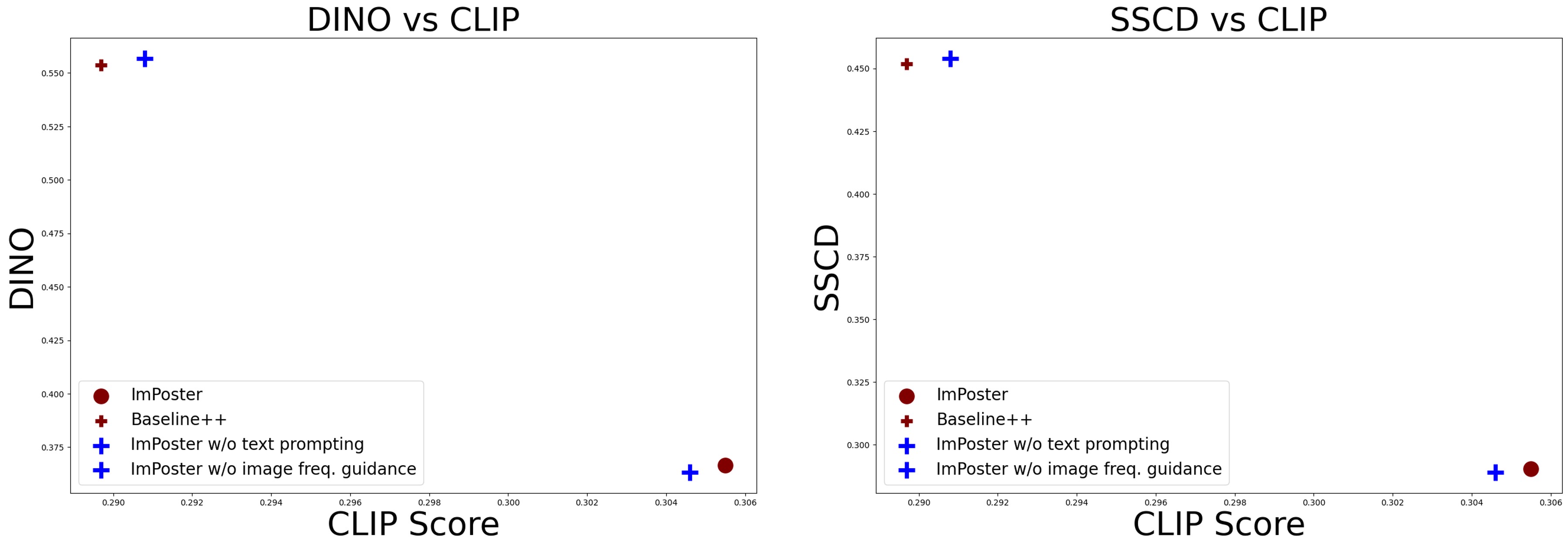}
    \caption{ImPoster is able to successfully transfer the driving motion while retaining the characteristics of the source subject, and achieves a better trade-off between driving action (CLIP/Phase score) and source subject (SSCD/DINO) than prior work, as also evidenced by our qualitative results. Stepwise text prompting creates a prior for the driving action to enable the model generate the driving action. The image frequency guidance formulations help the model in preserving the characteristics of the source subject, while reinforcing the driving action.}
    \label{fig:quant}
    \vspace{-2pt}
\end{figure*}

\begin{figure*}[h]
    \centering
    \includegraphics[scale=0.22]{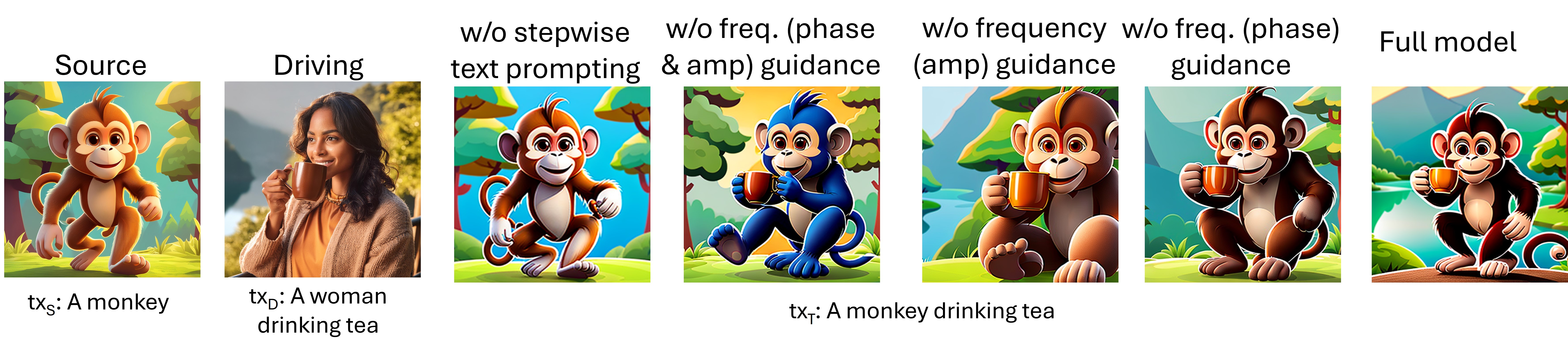}
    \caption{\textbf{Ablations.} Without stepwise text prompting, there is no prior for the driving action, which inhibits the model from generating the driving action accurately. The frequency (amplitude and phase) guidance methods help in generating the characteristics of the source subject (here, monkey) accurately - notice that there are changes to the color of the monkey (column 6), missing details in the fingers (column 4), changes in the size of the monkey (column 5). }
    \label{fig:ablation1}
    %\vspace{-25pt}
\end{figure*}

\paragraph{Quantitative Results: Comparisons with Baseline++ and Ablations}

In concordance with prior work~\cite{ruiz2022dreambooth,kawar2022imagic,kumari2022multi} on diffusion models for text-based image editing/ personalization, we evaluate our method using the following quantitative metrics. We compute the averages using all $600$ generated images. The results are in Figure~\ref{fig:quant}.  

\begin{enumerate}
    \item \textbf{Text alignment:} We evaluate the alignment of the generated target image with the target text using the CLIP score~\cite{ruiz2022dreambooth}. Higher CLIP scores indicate higher alignment with text. Here, CLIP Score gives us a broad overview of the alignment of the generated image with the subject and action. ImPoster achieves a far higher CLIP score than Baseline++, indicating its ability to effectively transfer the driving action to the source subject.   
    \item \textbf{Fidelity:} We evaluate the fidelity of the subject in the generated target image the source subject using self-supervised similarity metrics -  SSCD~\cite{pizzi2022self} and DINO scores~\cite{caron2021emerging}. Baseline++ is unable to generate the driving action, and simply replicates the source image. This results in it having a higher value of SSCD and DINO score than ImPoster, which is able to transfer the driving action, as well as preserve the characteristics of the source image. 
    \item \textbf{Phase score:} To evaluate the correctness of the action or pose generated in the target image (as dictated by the driving image), we define a new metric called the phase score. As per classical computer vision literature~\cite{oppenheim1981importance}, the phase of the Fourier transform of the image is indicative of the action depicted in the image. We compute phase score as the cosine similarity between the phase of the Fourier transform of the driving image and the generated target image. Higher cosine score indicates higher similarity. ImPoster achieves a higher phase score of $0.7529$ as compared to Baseline++'s phase score of $0.7515$, indicating its ability to transfer the driving action effectively. 
\end{enumerate}

\paragraph{Ablation analysis.} \textbf{a. Overall alignment with driving action and source subject:} ImPoster achieves a higher CLIP score than all ablations, indicating the usefulness of each component of our model towards transferring the driving action to the source subject, while preserving its characteristics. The stepwise text prompting and frequency guidance methods (including phase frequency guidance and amplitude frequency guidance) are complementary to each other and work in a holistic manner to achieve the desired goal. \textbf{b. Effectiveness of stepwise text prompting:} The stepwise text prompting strategy provides a crucial signal for generating the driving image, without which, the model achieves an overall low CLIP score as well as Phase score. Similar to Baseline++, the model without the stepwise text prompting strategy has a tendency to replicate the source image without the driving action, resulting in a higher SSCD or DINO score. \textbf{c. Effectiveness of image frequency guidance:} Without the frequency guidance functions, the SSCD and DINO scores drop, indicating its effectiveness in preserving fidelity w.r.t. source subject while transferring the driving action. 

\begin{table*}
    \centering
    \resizebox{0.99\linewidth}{!}{
    \begin{tabular}{ccccc}
    \toprule 
        Method & CLIP & Phase Score & DINO & SSCD \\
        \midrule 
        %\rowcolor{RowColorCode}
        %\multicolumn{8}{c}{Dataset: \model~- Syn} \\
        %\midrule
        %Baseline++ &  $0.2897$ &  $0.7515$ & $0.5538$ & $0.4517$ \\
        ImPoster & $0.3055$ & $0.7529$ & $0.3665$ & $0.2903$ \\
       \midrule 
       Effectiveness of Stepwise text prompting: set k=0 & $0.2908$ & $0.7516$ & $0.5568$ & $0.454$ \\
       Effectiveness of image frequency guidance: set $s_{a}=0, s_{p}=0$ & $0.3046$ & $0.7529$ & $0.3633$ & $0.2890$ \\
       Effectiveness of image frequency (phase) guid. : set $s_{p}=0$ & $0.3052$ & $0.7528$ & $0.3632$ & $2884$ \\
       Effectiveness of image frequency (amp) guid. : set $s_{a}=0$ & $0.3062$ & $0.7529$ & $0.3629$ & $2877$ \\
       \bottomrule
    \end{tabular}
    }
    %\vspace*{-0.5em}
    \caption{ImPoster is able to successfully transfer the driving motion while retaining the characteristics of the source subject, and achieves a better trade-off between driving action (CLIP/Phase score) and source subject (SSCD/DINO) than prior work, as also evidenced by our qualitative results. Stepwise text prompting creates a prior for the driving action to enable the model generate the driving action. The image frequency guidance formulations help the model in preserving the characteristics of the source subject, while reinforcing the driving action.}
    %\vspace*{-1.5em}
    \label{tab:metrics}
\end{table*}

%\paragraph{Small-scale user study}

%We conducted a small-scale user study to compare our method with the best baseline, Baseline++ (or multi-concept DreamBooth LoRA). For 12 random examples, we show participants 4 images - the source image, the driving image, result corresponding to Baseline++ and result corresponding to ImPoster. Each example is shown to 26 users who are asked to choose the better result between Baseline++ and ImPoster following the question: Please indicate the image which better corresponds to the 'driving' action, while preserving the identity of the 'source' character. Please note that we are looking to determine the image that is better in terms of holistically preserving both, 'driving' action as well as 'source' character. Our method was preferred by $69.96\%$, demonstrating that it ranks significantly higher than the best baseline in human evaluations.
\section{Conclusions}

We propose the usage of infusing text and image frequency for the task of method for the image generation task of subject-driven action personalization. To the best of our knowledge, ours is the first approach towards this task. Our method is successful in achieving a wide range of image animations such as making a monkey meditate and play the violin, as also validated by our quantitative results on our curated dataset using various metrics, including the newly proposed `phase score'. We hope our paper inspires further research in the area. 

\begin{figure*}[h]
    \centering
    \includegraphics[scale=0.3]{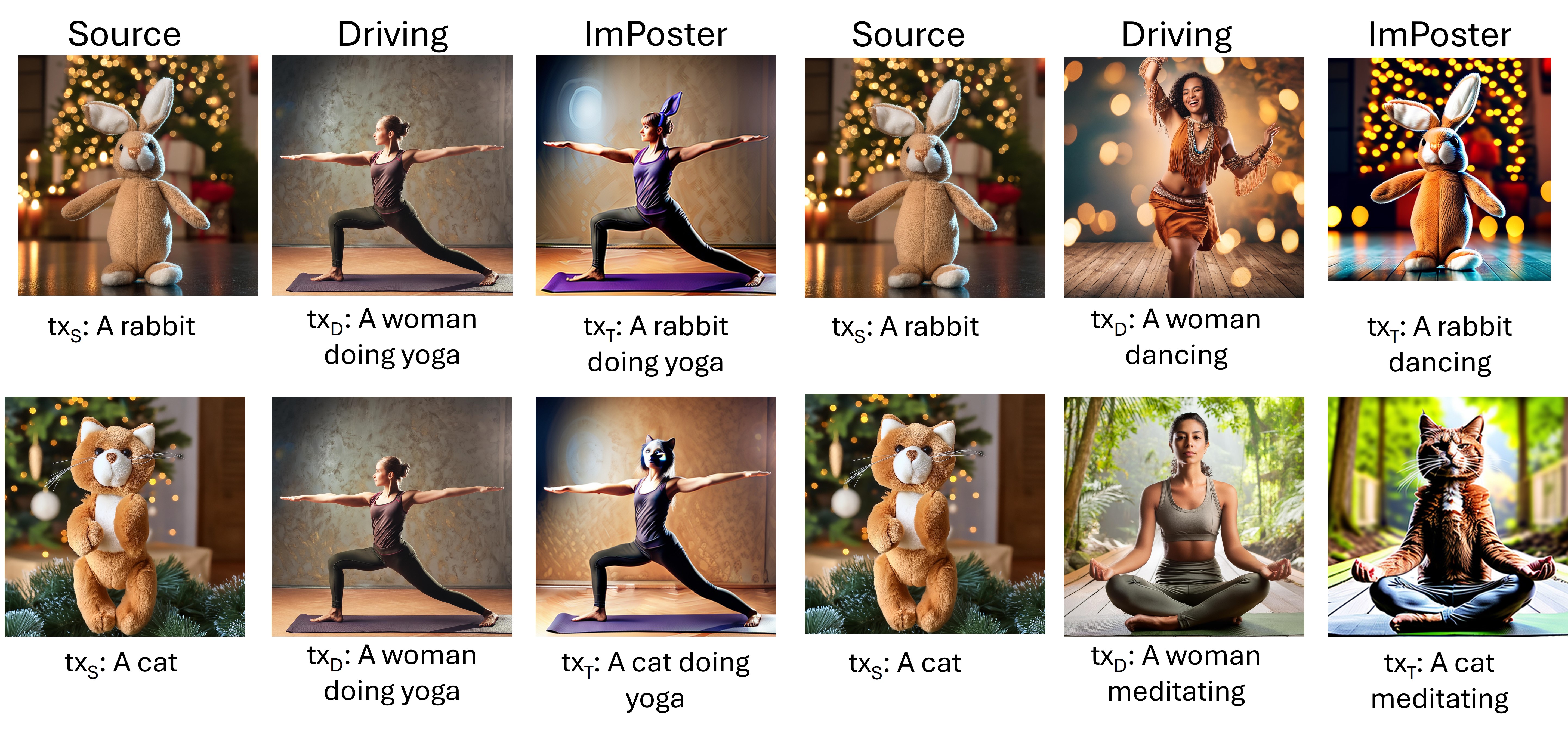}
    \caption{\textbf{Failure cases.} Our method is ineffective in a few cases - it is unable to make a rabbit perform yoga or dance, and a cat perform yoga or meditate. We believe that this is due to bias-variance trade-off issues, that remain unresolved even after the application of our stepwise text prompting and image frequency guidance strategies. Usage of stronger vision and language backbones in the text to image generation pipeline, as they are made open-source to the community, can help the model disentagle features of the image better to alleviate these bias-variance trade-off issues. Besides, further  research on this problem can lead to the development of newer methods that can help improve these results.}
    \label{fig:abltion1}
    %\vspace{-25pt}
\end{figure*}

\section{Limitations and Future Work}

While our method is able to execute large non-trivial non-rigid pose transfers to specific subjects, as defined by a driving image, it is still limited in the ability of change that it can bring forth. For instance, it is unable to make a rabbit or a cat do yoga, or deal with human subjects, indicating that there is scope for further research. Our proposed solution is a plug-and-play method. With the emergence of newer and stronger backbones, our method can be seamlessly integrated and utilized. Another direction for future research, also a widespread issue in the text-based image personalization literature, is the development of better quantitative metrics for comprehensively measuring subject fidelity and the desired edit. Since self-supervised methods such as SSCD and DINO measure image-level similarity, they result in a high value even if there is absolutely no action transfer -- this leads to inappropriate evaluation of the models' capability in performing the desired edit (driving action transfer in our case), while maintaining source subject fidelity. More directions for future work including investigating our frequency guidance strategy for other image editing and personalization applications, extension to video applications and scenarios that involve more than one source subject or driving action in the scene. 

\textbf{Societal impact.} Research on identification of fake imagery is essential to prevent the malicious use of our method.

\bibliography{references}
\bibliographystyle{iclr2025_conference}

\clearpage
\begin{figure*}
    \centering
    \includegraphics[scale=0.45]{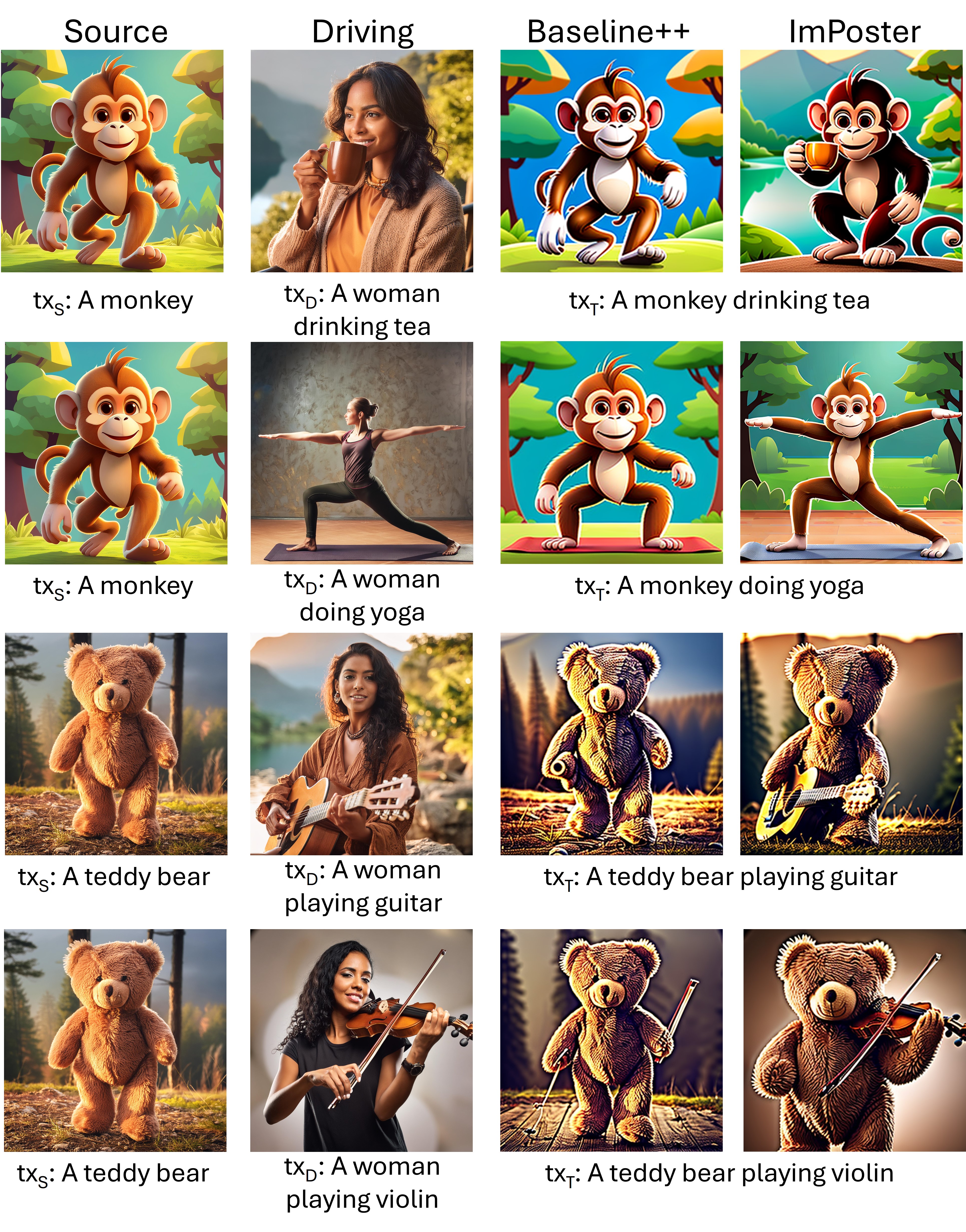}
    \caption{ImPoster is able to successfully transfer the driving action to a source subject, while maintaining its characteristics. In contrast, Baseline++ is unable to generate the driving action for the given source subject due to bias issues.}
    \label{fig:results2}
    %\vspace{-25pt}
\end{figure*}

\begin{figure*}
    \centering
    \includegraphics[scale=0.45]{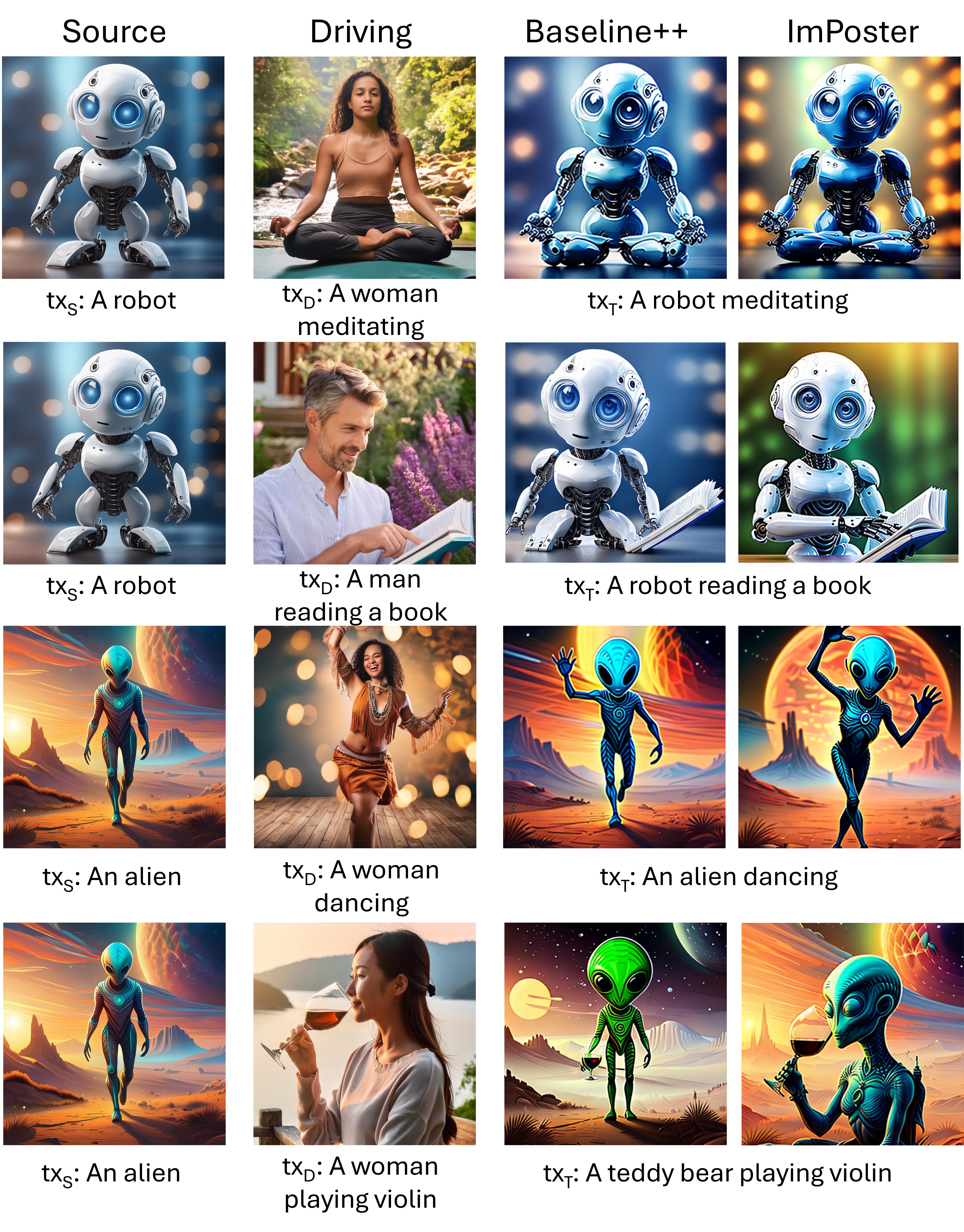}
    \caption{ImPoster is able to successfully transfer the driving action to a source subject, while maintaining its characteristics. In contrast, Baseline++ is unable to generate the driving action for the given source subject due to bias issues.}
    \label{fig:results3}
    %\vspace{-25pt}
\end{figure*}

\begin{figure*}
    \centering
    \includegraphics[scale=0.45]{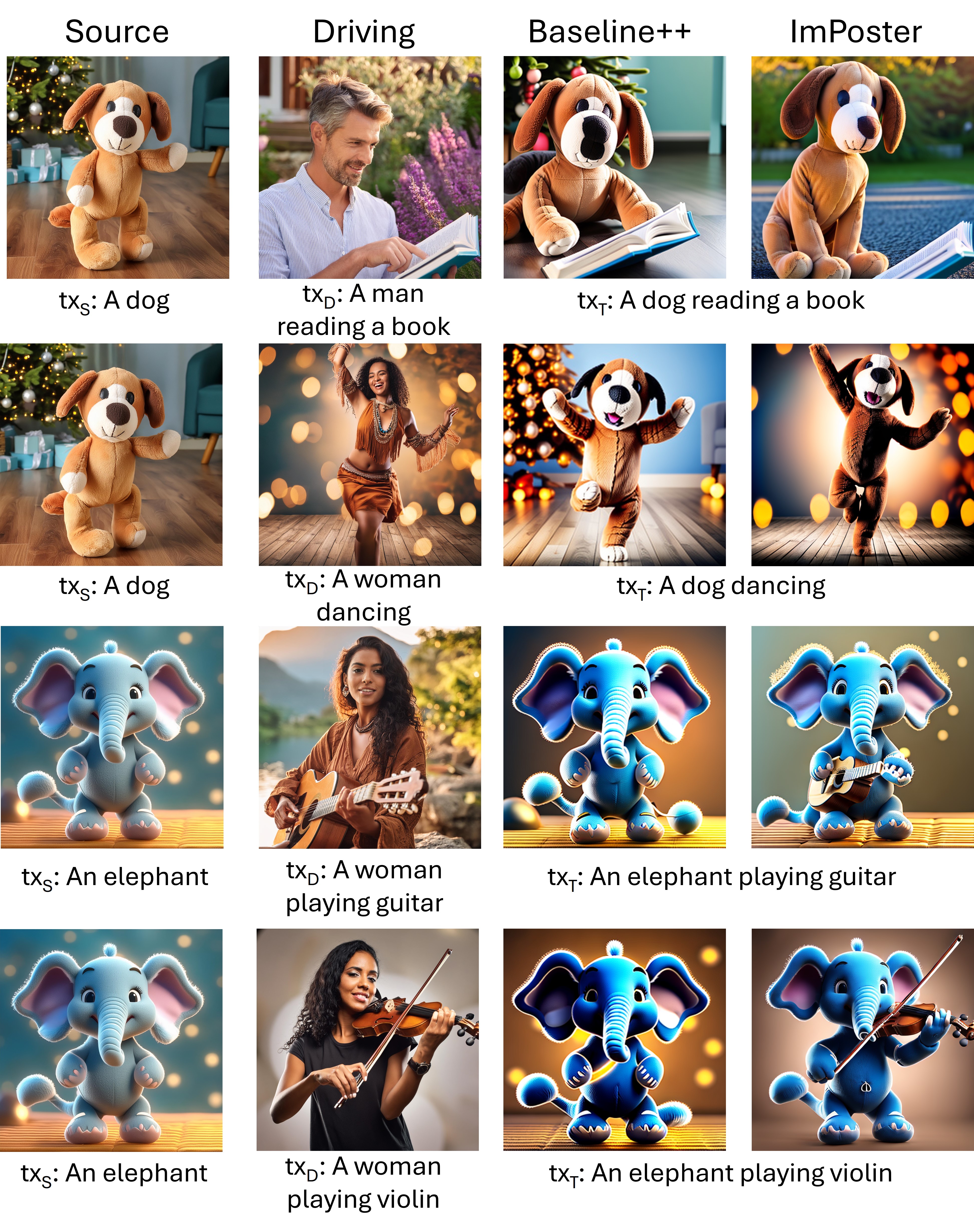}
    \caption{ImPoster is able to successfully transfer the driving action to a source subject, while maintaining its characteristics. In contrast, Baseline++ is unable to generate the driving action for the given source subject due to bias issues.}
    \label{fig:results4}
    %\vspace{-25pt}
\end{figure*}

\begin{figure*}
    \centering
    \includegraphics[scale=0.45]{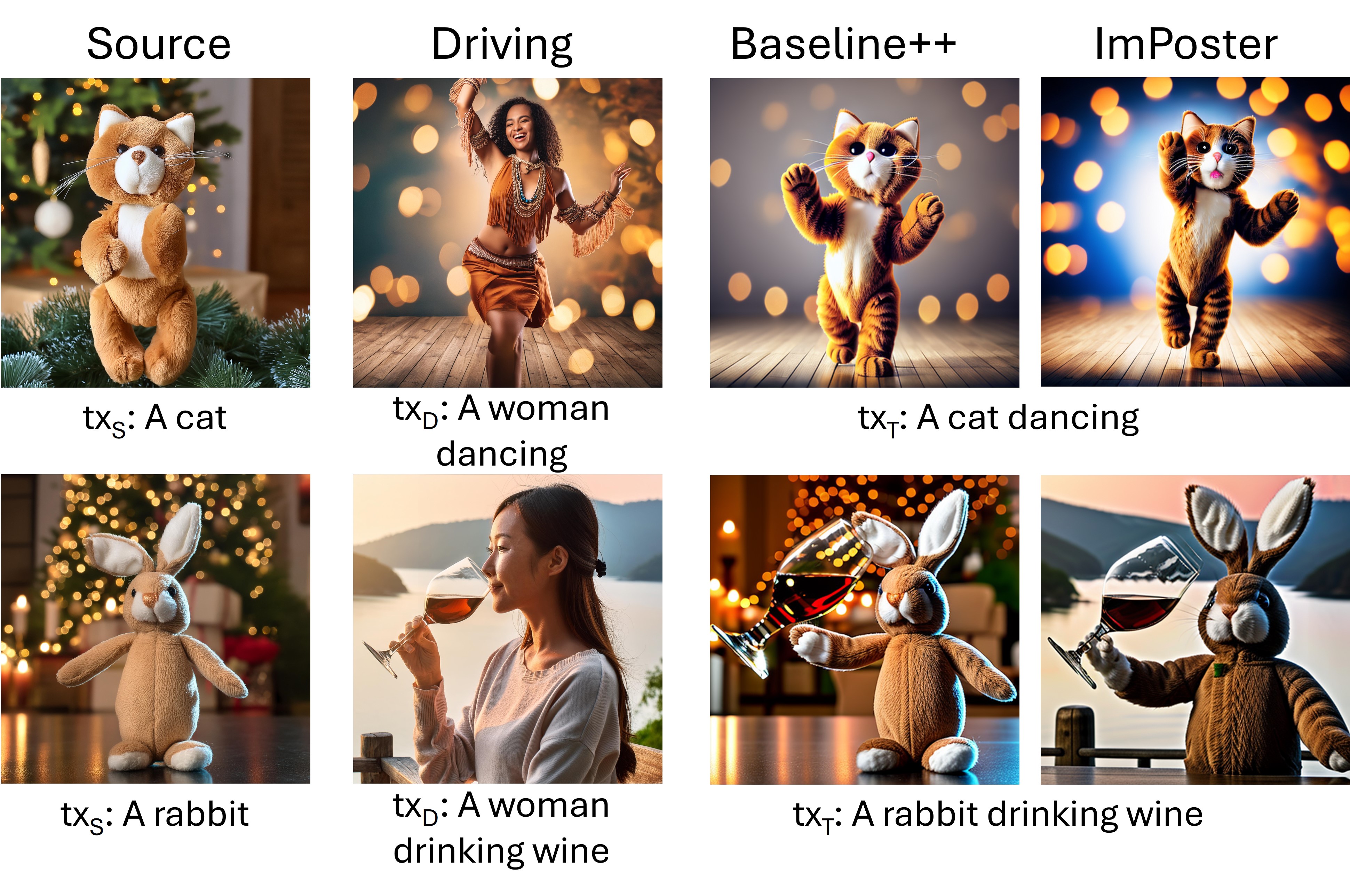}
    \caption{ImPoster is able to successfully transfer the driving action to a source subject, while maintaining its characteristics. In contrast, Baseline++ is unable to generate the driving action for the given source subject due to bias issues.}
    \label{fig:results5}
    %\vspace{-25pt}
\end{figure*}

\end{document}